# Advances in Probabilistic Reasoning

**Dan Geiger**
Northrop Research and Technology Center
One Research Park
Palos Verdes, CA 90274

**David Heckerman**
Departments of Computer Science and Pathology
University of Southern California
HMR 204, 2025 Zonal Ave, LA, CA 94305

## Abstract

This paper discuses multiple Bayesian networks representation paradigms for encoding asymmetric independence assertions. We offer three contributions: (1) an inference mechanism that makes explicit use of asymmetric independence to speed up computations, (2) a simplified definition of similarity networks and extensions of their theory, and (3) a generalized representation scheme that encodes more types of asymmetric independence assertions than do similarity networks.

## 1 Introduction

Traditional probabilistic approaches to diagnosis, classification, and pattern recognition face a critical choice: either specify precise relationships between all interacting variables or make uniform independence assumptions throughout. The first choice is computationally infeasible except in very small domains, while the second, which is rarely justified, often yields inadequate conclusions.

Bayesian networks offer a compromise between the two extremes by encoding independence when possible and dependence when necessary. They allow a wide spectrum of independence assertions to be considered by the model builder so that a practical balance can be established between computational needs and adequacy of conclusions.

Although Bayesian networks considerably extend traditional approaches, they are still not expressive enough to encode every piece of information that might reduce computations. The most obvious omissions are *asymmetric independence* assertions stating that variables are independent for some but not necessarily for all of their values. Such asymmetric assertions cannot be represented naturally in a Bayesian network. Several researchers observed this limitation, however, until recently no effort was made to remove it.

Similarity network paradigm is the first major effort towards the representation of asymmetric independence [Heckerman, 1990]. Contingent influence diagrams is an alternative approach [Fung and Shachter, 1991]. Both schemes employ asymmetric independence to ease the elicitation and improve the quality of probabilistic models.

This article offers three contributions: (1) an inference mechanism that makes explicit use of asymmetric independence to speed up computations, (2) a simplified definition of similarity networks and extensions of their theory, and (3) a generalized representation scheme that encodes more types of asymmetric independence assertions than do similarity networks.

These contributions address problems of knowledge representation, inference, and knowledge acquisition. In particular, Section 2 describes *Bayesian multinets* and how to use them for inference, Section 3 describes knowledge acquisition using *similarity networks* and how to convert them to Bayesian multinets, Section 4 extends these representation schemes to the case where hypotheses are not mutually exclusive and section 5 summarizes the results. We assume the reader is familiar with the definition and usage of Bayesian networks. For details consult [Pearl, 1988].

## 2 Representation and Inference

### 2.1 Bayesian Multinets

The following example demonstrates the problem of representing asymmetric independence by Bayesian networks:

> A guard of a secured building expects three types of persons to approach the building's entrance: workers in the building, approved visitors, and spies. As a person approaches the building, the guard notes its gender and whether or not the person wears a badge. Spies are mostly men. Spies always wear badges in order to fool the guard. Visitors don't wear badges because they don't have one. Female-workers tend to wear badges more often than do male-workers. The task of the guard is to identify the type of person approaching the building.

A Bayesian network that represents this story is shown in Figure 1. Variable $h$ in the figure represents the correct identification. It has three values $w$, $v$, and $s$ respectively denoting worker, visitor, and spy. Variables $g$ and $b$ are binary variables representing, respectively, the person's gender and whether or not the person wears a badge. The links from $h$ to $g$ and from $h$ to $b$ reflect the fact that both gender and badge-wearing are clues for correct identification, and the link from $g$ to $b$ encodes the relationship between gender and badge-wearing.

Unfortunately, the topology of this network hides the fact that, independent of gender, spies always wear badges and visitors never do. The network does not show that gender and badge-wearing are conditionally independent given the person is a spy or a visitor. A link between $g$ and $b$ is drawn merely because gender and badge-wearing are related variables when the person is a worker.

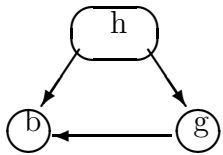

Figure 1: A Bayesian network for the secured-building example.

We can more adequately represent this story using two Bayesian networks shown in Figure 2. The first network represents the cases where the person approaching the entrance is either a spy or a visitor. In these cases, badge-wearing depends merely on the type of person approaching, not on its gender. Consequently, nodes $b$ and $g$ are shown to be conditionally independent (node $h$ blocks the path between them). The links from $h$ to $b$ and from $h$ to $g$ in this network reflect the fact that badges and gender are relevant clues for distinguishing between spies and visitors. The second network represents the hypothesis that the person is a worker, in which case gender and badge-wearing are related as shown.

Figure 2 is a better representation than Figure 1 because it shows the dependence of badge-wearing on gender only in context in which such a relationship exists, namely, for workers. Moreover, the former representation requires 11 parameters while the representation of Figure 2 requires only 9. This gain, due to asymmetric independence, could be substantially larger for real-sized problems because the number of parameters needed grows exponentially in the number of variables, whereas the overhead of representing multiple networks grows only linearly.

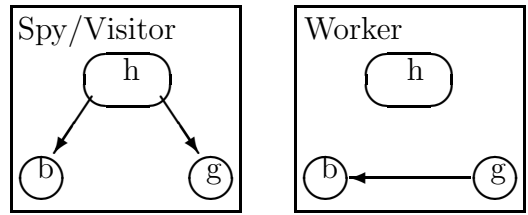

Figure 2: A Bayesian multinet representation of the secured-building story.

We call the representation scheme of figure 2, a *Bayesian multinet*.

**Definition** Let $\{u_1 \ldots u_n\}$ be a finite set of variables each having a finite set of values, $P$ be a probability distribution having the Cartesian product of these sets of values as its sample space, and $h$ be a distinguished variable among the $u_i$'s that represents a mutually-exclusive and exhaustive set of hypotheses. Let $A_1, ..., A_k$ be a partition of the values of $h$. A directed acyclic graph $D_i$ is called a *local network* of $P$ (associated with $A_i$) if it is a Bayesian network of $P$ given that one of the hypotheses in $A_i$ holds, i.e., $D_i$ is a Bayesian network of $P(u_1 \ldots u_n | A_i)$. The set of $k$ local networks is called a *Bayesian multinet* of $P$.[1]

In the secured-building example of Figure 2, $\{\{spy, visitor\}, \{worker\}\}$ is a partition of the values of the hypothesis node $h$, one local network is a Bayesian network of $P(h, b, g | worker)$ and the other local network is a Bayesian network of $P(h, b, g | \{\text{spy, visitor}\})$. [2]

The fundamental idea of multinets is that of *conditioning*; each local network represents a distinct situation conditioned that hypotheses are restricted to a specified subset. Savings in computations and space occur because, as a result of conditioning, asymmetric independence assertions are encoded in the topology of the local networks. In the example above, conditional independence between gender and badge-wearing is encoded as a result of conditioning on $h$.

Notably, conditioning may also destroy independence relationships rather then create them [Pearl, 1988].

---

[1] A Bayesian multinet roughly corresponds to an *hypothesis-specific similarity network* as defined in Heckerman's dissertation (1990, page 76).

[2] The conditioning set $\{spy, visitor\}$ is a short hand notation for saying that $h$ draws its values from this set, namely, either $h = spy$ or $h = visitor$.

However, if the distinguished variable is a root node (i.e., a node with no incoming links), conditioning on its values never decreases and often increases the number of independence relationships, resulting in a more expressive graphical representation. Other situations are addressed below where the hypothesis variable is not a root node or where more than one node represents hypotheses.

## 2.2 Representational and Computational Advantages

The vanishing dependence between gender and badge-wearing is an example of an *hypothesis-specific* independence because it is manifest only when conditioning on specific hypotheses, that is, for spies and visitors, but not for workers. The following variation of the secured-building example demonstrates an additional type of asymmetric independence that can be represented by Bayesian multinets as well.

> The guard of the secured building now expects *four* types of persons to approach the building's entrance: executives, regular workers, approved visitors, and spies. The guard notes gender, badge-wearing, and whether or not the person arrives in a limousine ($l$). We assume that only executives arrive in limousines and that male and female executives wear badges just as do regular workers (to serve as role models).

This story is represented by the two local networks shown in Figure 3. One network represents a situation where either a spy or a visitor approaches the building, and the other network represents a situation where either a worker or an executive approaches the building. The link from $h$ to $l$ in the latter network reflects the fact that arriving in limousines is a relevant clue for distinguishing between workers and executives. The absence of this link in the former network reflects the fact that it is not relevant for distinguishing between spies and visitors.

The vanishing dependence between gender and the hypothesis variable $h$ when $h$ is restricted to a subset of hypotheses {*worker, executive*} is an example of *subset independence*. Similarly, badge-wearing is independent of $h$ when restricted to {*worker, executive*}, and arriving in limousines is independent of $h$ when restricted to {*spy, visitor*}. [3]

Subset independence is a source of considerable computational savings. For example, in lymph-node pathology less than 20% of the potential morphological findings are relevant for distinguishing any given pair of disease hypotheses (among over 60 diseases) [Heckerman, 1990].

---
[3] Heckerman coined the terms subset independence and hypothesis-specific independence in his dissertation.

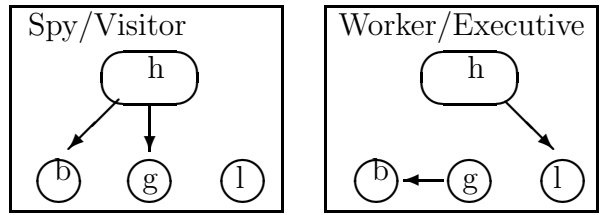

Figure 3: A Bayesian multinet representation of the augmented secured-building story.

Below we demonstrate these computational savings using the simple secured-building example; more savings are obtained in real domains such as lymph-node pathology.

Suppose the guard sees a male ($\mathbf{g}$) wearing a badge ($\mathbf{b}$) approaches the building and suppose the guard doesn't notice whether or not the person arrives in a limousine. A computation of the posterior probability of each possible identification (*executive, worker, visitor, spy*) based on the Bayesian network of Figure 1 simply yields the chaining rule:

$$P(h|\mathbf{g}, \mathbf{b}) = K \cdot P(h) \cdot P(\mathbf{g}|h) \cdot P(\mathbf{b}|\mathbf{g}, h). \quad (1)$$

where $K$ is the normalizing constant.

Using the representation of Figure 3, however, the following more efficient computations are done instead:

$$P(spy|\mathbf{g}, \mathbf{b}) = K \cdot P(spy) \cdot P(\mathbf{g}|spy) \cdot P(\mathbf{b}|spy) \quad (2)$$
$$P(visitor|\mathbf{g}, \mathbf{b}) = K \cdot P(visitor) \cdot P(\mathbf{g}|visitor) \cdot$$
$$P(\mathbf{b}|visitor) \quad (3)$$
$$P(worker|\mathbf{g}, \mathbf{b}) = K \cdot P(worker) \cdot P(\mathbf{g}|worker) \cdot$$
$$P(\mathbf{b}|\mathbf{g}, worker) \quad (4)$$
$$P(\mathbf{g}, \mathbf{b}|executive) = P(\mathbf{g}, \mathbf{b}|worker). \quad (5)$$

Equations 2 and 3 take advantage of an hypothesis-specific independence assertion, namely, that $g$ and $b$ are conditionally independent given, respectively, that $h = spy$ and $h = visitor$. Equation 5 uses a subset independence assertion, namely, that $b$ and $g$ are independent of $h$ restricted to {*worker, executive*}.

More generally, calculating the posterior probability of each hypothesis based on a set of observations $e_1, ..., e_m$ is done in two steps. First, for each hypothesis $h_i$, the probability $P(e_1, ..., e_m|h_i)$ is computed via standard algorithms such as Spiegelhalter and Lauritzen's (88) or Pearl's (88). Second, these results are combined via Bayes' rule:

$$P(h_i|e_1...e_m) = K \cdot p(h_i) P(e_1...e_k|h_i). \quad (6)$$

Notably, the computation of $P(e_1 \ldots e_k|h_i)$ in the first step uses the local networks as done in Eqs. (2) through

(5) and does not use a single Bayesian network as done in Eq. (1). Consequently, when the values of $h$ are properly partitioned, the extra independence relationships encoded in each local network could considerably reduce computations.

The parameters needed to perform the above computations consist, as we shall see next, of the prior of each hypothesis $h_i$ and the parameters encoded in the local networks:

**Theorem 1** *Let $\{u_1 \ldots u_n\}$ be a finite set of variables each having a finite set of values, $P$ be a probability distribution having the Cartesian product of these sets of values as its sample space, $h$ be a distinguished variable among the $u_i$s, and $M$ be a Bayesian multinet of $P$. Then, the posterior probability of every hypothesis given any value combination for the variables in $\{u_1 \ldots u_n\}$ can be computed from the prior probability of $h$'s values and from the parameters encoded in $M$.*

According to Eq. 6 above, the only parameters needed for computing the posterior probability of each hypothesis $h_i$, aside of the priors, are $p(v_2 \ldots v_n | h_i)$ where $v_2 \ldots v_n$ are arbitrary values of $u_2 \ldots u_n$ (assuming without loss of generality that $h = u_1$). Let $D_i$ denote a local network in $M$, $A_i$ be the hypotheses associated with $D_i$, and $h_i$ be an hypothesis in $A_i$. Clearly, $p(v_2 \ldots v_n | h_i)$ is equal to $p(v_2 \ldots v_n | h_i, A_i)$ because $h_i$ logically implies the disjunction over all hypotheses in $A_i$. The latter probability is computable from the local network $D_i$ by any standard algorithm (e.g., [Pearl, 1988]), thus, the former is also computable as needed. □

For example, $P(\mathbf{g}|worker, \{worker, executive\})$ is equal to the probability $P(\mathbf{g}|worker)$ because $worker$ logically implies the disjunction $worker \vee executive$. In fact, $P(\mathbf{g}|worker, \{worker, executive\})$ is also equal to $P(\mathbf{g}|\{worker, executive\})$ because $\mathbf{g}$ and $worker$ are independent given $\{worker, executive\}$ as shown in Figure 3. In this example, the needed probability $P(\mathbf{g}|worker)$ is equal to the given one $P(\mathbf{g}|\{worker, executive\})$, however in general, the needed probabilities are computed via standard inference algorithms.

### 2.3 Overcoming some Limitations

The multinet approach described thus far is especially beneficial when the hypothesis variable can be modeled as a root node because, then, no dependencies are ever introduced by conditioning on the different hypotheses. However, the hypothesis node cannot always be modeled as a root node. For example, in the secured-building story, suppose there are two independent reports indicating possible spying, say, for military and economical reasons respectively. Such a priori factors for correct identification are modeled as parent nodes of $h$, called, say, *economics* and *military* having no link between them to show their mutual independence. The resulting network in this case is simply $economics \rightarrow h \leftarrow military$.

However when $h$ assumes the value *spy*, an induced link is introduced between its parents *economics* and *military*; one explanation for seeing a spy changes the plausibility of the other explanation, thus making the two variables economics and military be not independent conditioned on $h = spy$. Consequently, an induced link must be drawn between the *economics* and *military* nodes in the local network for spies vs. visitors to account for the above dependency. This link would not appear in the full Bayesian network because economics and military are marginally independent (they become dependent only when conditioning on $h = spy$). Such induced links are often hard to quantify and therefore, constructing a single local network is sometimes harder than constructing the full network, as is the case in the above example.

One approach to handle this situation is to first construct a Bayesian network that represents only a priori factors that influence the hypotheses, ignoring any evidential variables (such as gender, badge-wearing, and limousines). In our example, this network would be $economics \rightarrow h \leftarrow military$. Then, use this network to revise the a priori probabilities of the different hypotheses. Finally, construct local networks ignoring a priori factors (as done in Figure 2) and use the resulting multinet with the revised priors of $h$ to compute the posterior probability of $h$ as determined by the evidential clues. This decomposition technique works best if a priori factors are independent of all clues conditioned on the different hypotheses. That is, in situations that can be modeled with Bayesian networks of the form shown in Figure 4 where all paths between a priori factors $r_i$'s and evidential clues $f_i$'s pass through $h$.

When a network of this form cannot serve as a justifiable model, another approach can be used instead; compose a Bayesian multinet ignoring a priori factors, construct a Bayesian network from the local networks by taking the union of all their links (e.g., the union of all links in Figure 2 yields the Bayesian network of Figure 1). Finally, add a priori factors to the resulting network. This approach was proposed in [Heckerman, 1990].

The disadvantage of this method is that in the process of generating a Bayesian network from a multinet, one encodes asymmetric independence in the parameters rather than in the topology of the Bayesian network. Consequently, these asymmetric assertions are not available to standard inference algorithm to speed up their computations.

Nevertheless, this approach is still the best alternative for decomposing the construction of large Bayesian networks having topologies more complex than that of Figure 4. Such decomposition techniques are crucially needed due to the overwhelming details of real-life problems. Additional issues of knowledge acquisition are discussed below.

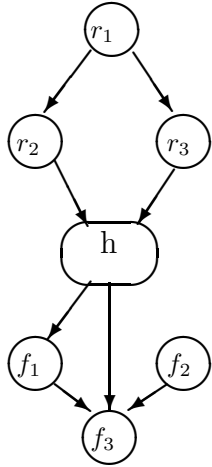

Figure 4: A Bayesian network where all paths between a priori factors $r_i$'s and evidential clues $f_i$'s pass through $h$.

## 3 Knowledge Acquisition/ Representation

### 3.1 Similarity Networks

Recall the guard that must distinguish between workers, executives, visitors and spies. In this story, some variables do not help distinguish between certain hypotheses. For example, gender and badges do not help distinguish between workers and executives, and limousines do not help distinguish between spies and visitors. In richer domains, large numbers of variables are often not relevant for distinguishing between certain hypotheses.

Unfortunately, the Bayesian multinet approach requires full specification of all variables in each local network even when they are not relevant to distinguish between the hypotheses associated with that local network. For example the relationship between $b$ and $g$ is encoded in the local network for spies vs. visitors although these variables do not help distinguish between this pair of hypotheses (Figure 3). Assessing such relationships, in contexts where they are not relevant, poses insurmountable burden on the expert consulted as is demonstrated by the following quote [Heckerman, 1990]:

> "When the expert pathologist was asked questions of the form
>
>> Given any disease, does observing feature $x$ change your belief that you will observe feature $y$ ?
>
> the expert sometimes would reply
>
>> I've never thought about these two features at the same time before. Feature $x$ is relevant to only one set of diseases, while feature $y$ is only relevant to another set of diseases. These sets of diseases do not overlap, and I never confuse the first set of diseases with the second."

The solution is to simply include in each local network only those variables that are relevant for distinguishing between the hypothesis covered by that local network.

However, by doing so, valuable information for correct identification might be lost. For example, the relationships between badge-wearing and gender in Figure 3 would be lost. To compensate for such losses of information, additional local networks must be constructed.

For example, the secured-building can be represented with three local networks shown in Figure 5 rather than two as in Figure 3. One network is used to distinguish between spies and visitors, another between visitors and workers, and a third between workers and executives. In each local network we include only those variables relevant to distinguishing the hypotheses covered by that local network. In particular, the relationship between badge-wearing and gender is not included in the local network for workers vs. executives as in Figure 3. This relationship, however, is included in the local networks for visitors vs. workers because it helps distinguish between these two hypotheses. The reason for not loosing needed information is that the three local networks are based on a *connected cover* of hypotheses (rather than a partition).

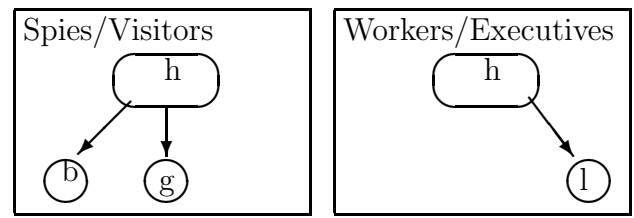

Figure 5: A similarity network representation of the secured-building story.

**Definition** A *cover* of a set $A$ is a collection $\{A_1, ..., A_k\}$ of non-empty subsets of $A$ whose union is $A$. Each cover is a hypergraph, called the *similarity hypergraph*, where the $A_i$'s are edges and elements of $A$ are nodes. A cover is *connected* if the similarity hypergraph is connected.

In Figure 5, {*spy, visitor*}, {*visitor, worker*}, {*worker, executive*} is a cover of the hypotheses set. This cover is connected because it is simply a four-nodes chain *spy—visitor—worker—executive* which, by definition, is a connected hypergraph. The set {{*spy, visitor*}, {*worker, executive*}} is also a cover but it is not connected. The set {{*worker, executive, visitor*}, {*visitor, spy*}} is an example of a connected cover that is a hypergraph which is not a graph.

**Definition**  Let $U = \{u_1 \ldots u_n\}$ be a finite set of variables each having a finite set of values, $P$ be a probability distribution having the cross product of these sets of values as its sample space, and $h$ be a distinguished variable among the $u_i$'s that represents a mutually-exclusive and exhaustive set of hypotheses. Let $A_1, ..., A_k$ be a connected cover of the values of $h$. A directed acyclic graph $D_i$ is called a *comprehensive local network* of $P$ (associated with $A_i$) if it is a Bayesian network of $P$ assuming one of the hypotheses in $A_i$ holds, i.e., $D_i$ is a Bayesian network of $P(u_1 \ldots u_n | A_i)$. The network obtained from $D_i$ by removing nodes that are not relevant to distinguishing between hypotheses in $A_i$ is called an *ordinary local network*. The set of $k$ ordinary local networks is called an *(ordinary) similarity network* of $P$.

For example, the local networks of Figure 5 are ordinary, and together form an ordinary similarity network. Notably, hypotheses covered by each local network are often similar (e.g., spies and visitors), [4] a choice that maximizes the number of asymmetric independence relationships encoded.

Heckerman (1990) shows that under several assumptions, if a cover is connected, one can always remove from each local network variables that do not help distinguish between hypotheses covered by that local network and yet not loose the information necessary for representing the full joint distribution. These assumptions consist of 1) the hypothesis variable is a root node, 2) the cover is a graph and not a hypergraph, 3) the local networks are constrained by the same partial order, and 4) the distribution is strictly positive. Theses assumptions are relaxed below.

**Theorem 2**  *Let $\{u_1 \ldots u_n\}$ be a finite set of variables each having a finite set of values, $P$ be a probability distribution having the Cartesian product of these sets of values as its sample space, $h$ be a distinguished variable among the $u_i$s, and $S$ be a similarity network of $P$. Then, the posterior probability of every hypothesis given any value combination for the variables in $\{u_1 \ldots u_n\}$ can be computed from the parameters encoded in $S$ provided $p(h_i) \neq 0$ for every value $h_i$ of $h$.*

---

[4] Hence the name: similarity network.

To prove the above theorem, it suffices to consider the case where $h$ is a root node in all the local networks of $S$ because, otherwise, *arc-reversal* transformations [Shachter 1986] can be applied until $h$ becomes one.

Also note that since the similarity hypergraph is connected, it imposes $n-1$ independent equations among the following $n$: $p(h_i) = p(h_i|A_i) \cdot \sum_{h_j \in A_i} p(h_j)$, $i = 1 \ldots n$. In addition, $\sum_1^n p(h_i) = 1$. The values for $p(h_i)$ are the unique solution of these linear equations provided $p(h_i) \neq 0$ for $i = 1 \ldots n$.

Aside of the priors, the only remaining parameters needed for computing the posterior probability of each hypothesis $h_i$, are $p(v_2 \ldots v_n | h_i)$ where $v_2 \ldots v_n$ are arbitrary values of $u_2 \ldots u_n$ (assuming without loss of generality that $h = u_1$). Due to the chaining rule, $p(v_2 \ldots v_n | h_i)$ can be factored as follows:

$$p(v_2 \ldots v_n | h_i) = P(v_2|h_i) \cdot P(v_3|v_2 h_i) \ldots$$
$$p(v_n|v_1 \ldots v_{n-1} h_i).$$

Thus, it suffices to show that for each variable $u_j$, $p(v_j|v_2 \ldots v_{j-1} h_i)$ can be computed from the parameters encoded in $S$.

Let $D_i$ denote a local network in $S$, $A_i$ be the hypotheses associated with $D_i$, and $h_i$ be an hypothesis in $A_i$. There are two cases; either $u_j$ is depicted in $D_i$ or it is not. Let $A_i, A_{i+1} \ldots A_m$ be a path in the similarity hypergraph where $A_m$ is the only edge on this path associated with a local network that depicts $u_j$ as a node. If $u_j$ is depicted in $D_i$, then the path consists of one edge $A_i$ which is equal to $A_m$. If $u_j$ is not depicted in any local network, then $u_j$ does not alter the posterior probability of any hypothesis and is therefore omitted from the computations.

Let $D_k$ be the local netowrk associated with $A_k$ for $k = i+1 \ldots m$ and let $h_{i+1}, h_{i+2} \ldots h_m$ be a sequence of hypotheses such that $h_k \in A_{k-1} \cap A_k$. Due to the definition of similarity networks, since $u_j$ is not depicted in $D_k$ where $k < m$, the following equality must hold:

$$p(v_j|v_2 \ldots v_{j-1} h_{k-1}) = p(v_j|v_2 \ldots v_{j-1} h_k).$$

Since this equation holds for every $k$ between $i+1$ and $m$, we obtain,

$$p(v_j|v_2 \ldots v_{j-1} h_i) = p(v_j|v_2 \ldots v_{j-1} h_m).$$

Moreover,

$$p(v_j|v_2 \ldots v_{j-1} h_m) = p(v_j|v'_1 \ldots v'_l h_m)$$

where $u'_1 \ldots u'_l$ are the variables depicted in $D_m$ (a subset of $\{u_2 \ldots u_{j-1}\}$) because, due to the definition of similarity network, the variables deleted are conditionally independent of $v_j$, given the other variables; they are disconnected from all the other variables in $D_m$.[5]

Finally,
$$p(v_j|v'_1 \ldots v'_l h_m) = p(v_j|v'_1 \ldots v'_l h_m, A_m),$$
because $h_m$ logically implies the disjunction over all hypotheses in $A_m$.

The latter probability is computable from the local network $D_m$ by any standard algorithm (e.g., [Pearl, 1988]), thus, due the three equalities above, $p(v_j|v_2 \ldots v_{j-1} h_i)$ is also computable as needed. □

For example, to compute $P(g, b, l|spy)$ we use the following two equalities implied by Figure 5: From the first local network, $P(g, b, l|spy) = P(g|spy) \cdot P(b|spy) \cdot P(l|spy)$ and from the absence of $l$ in the first and second local networks, $P(l|spy) = P(l|worker)$. Thus, $P(g, b, l|spy) = P(g|spy) \cdot P(b|spy) \cdot P(l|worker)$, where all the needed probabilities are encoded in the similarity network. In fact, the proof of Theorem 2 provides a general way of factoring any desired probability, thus, the full joint distribution $P(g, b, l, h)$ is encoded in the ordinary similarity network of Figure 5.

Similarity networks have another important advantage not mentioned so far: protecting the model builder from omitting relevant clues. For example, suppose workers and executives often arrive with a smile to work (because the secured building is such a great place to be in) while spies and visitors arrive seriously. Such a clue, smile, is likely to be forgotten when constructing the local networks for spies vs. visitors and for visitors vs. executives because it does not help distinguish between these pairs of hypotheses. However, when constructing the similarity network of Figure 5, which includes a local network for distinguishing visitors from workers, smile is more likely to be recalled because the distinctions between visitors and workers are explicitly in focus.

### 3.2 Redundancy

Basing the construction of local networks on covers of hypotheses raises the problem of *redundancy*, namely, that some parameters are specified in more than one local network. For example, in Figure 5, the parameter $P(\mathbf{g}|visitor)$ should, in principle, be specified both in the first and in the second local network. This problem is particularly crucial because local networks are actually constructed from expert's judgments rather than from a coherent probability distribution as implied by the definition of similarity networks.

One way to remove redundancy is to automatically-translate a similarity network as it is being constructed to a Bayesian multinet which is never redundant. For example, instead of storing Figure 5, we can actually store Figure 3 which contains no redundant information.

---

[5]Geiger and Heckerman (1990) discuss weaker definitions of being irrelevant other than being disconnected.

The translation is done by the following algorithm.

#### Conversion Algorithm

**Input:** A similarity network $S$ of a probability distribution $P$.

**Output:** A Bayesian multinet of $P$.

1. For each ordinary local network $L$ in $S$:
   - Add a node for each variable not represented in $L$.
   - For each added node $x$, set the parents of $x$ in $L$ to be the union of all parents of $x$ in all other local networks where $x$ originally appeared, excluding variables that were originally in $L$.

2. Remove enough local networks from $S$ and enough hypotheses from the remaining local networks until a Bayesian multinet is obtained.

(A finer version of this algorithm is forthcoming).

Notably, the user of a similarity network need not know about the conversion to a Bayesian multinet which can be thought of as an internal representation. The user benefits from both the advantages of similarity network for knowledge acquisition, and from an inference algorithm (Section 2) that uses the Bayesian multinet produced by the conversion algorithm.

## 4 Generalized Similarity Networks

Previous sections assume all hypotheses are mutually exclusive and are, therefore, represented as values of a single hypothesis variable denoted $h$. Here this assumption is relaxed. We allow several variables to represent hypotheses, as needed by the following example:

> Consider the guard of Section 2 who has to distinguish between workers, visitors, and spies. A *pair* of people approach the building and the guard tries to classify them as they approach. Assume that only workers converse ($c$) and that workers often arrive with other workers (because they must car-pool to conserve energy).

A Bayesian network representing this situation is shown in Figure 6 where nodes $h_1$ and $h_2$ stand for the respective identity of the two persons. (The direction of the link between $h_1$ and $h_2$ is arbitrary.)

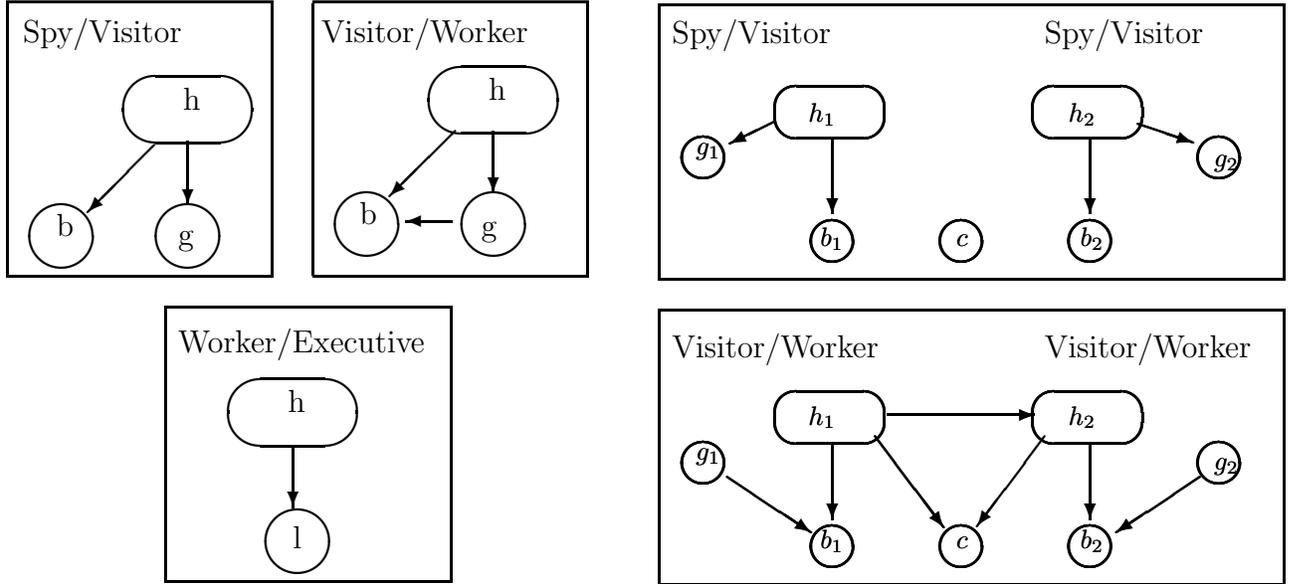

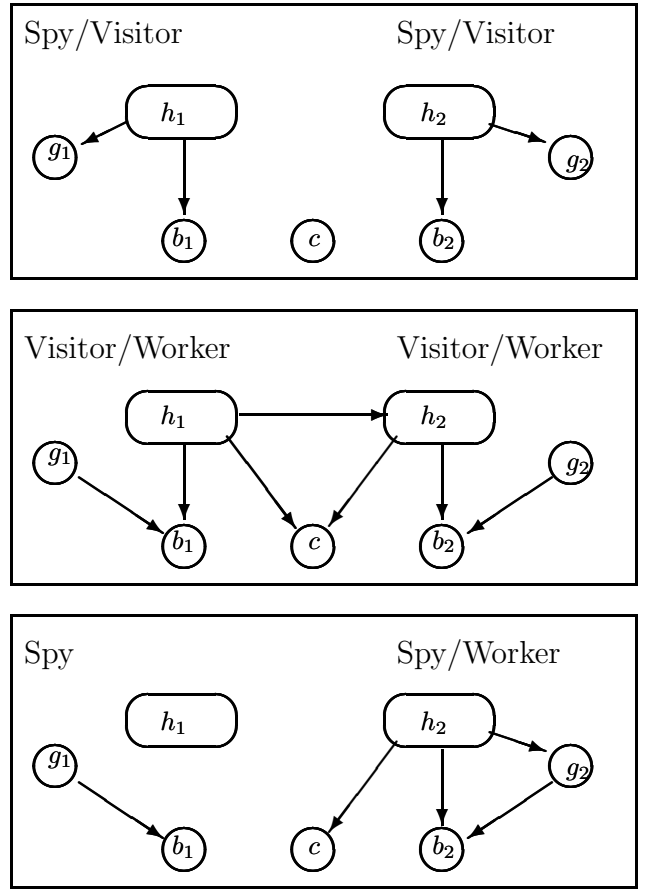

Figure 6: A Bayesian network with two hypothesis nodes $h_1$ and $h_2$.

Alternatively, we can represent this example using a *generalized similarity network*, or a *generalized Bayesian multinet*.

**Definition** Let $\{u_1 \ldots u_n\}$ be a finite set of variables each having a finite set of values, $P$ be a probability distribution having the cross product of these sets of values as its sample space, and $H$ be a subset of distinguished variables among the $u_i$'s each representing a set of hypotheses. Denote the Cartesian product of the sets of values of the distinguished variables by *domain(H)*. Let $A_1, ..., A_k$ be a connected cover of *domain(H)*. A directed acyclic graph $D_i$ is called a *comprehensive local network* of $P$ if it is a Bayesian network of $P(u_1 \ldots u_n | A_i)$. The network obtained from $D_i$ by removing nodes that are not relevant to distinguishing between hypotheses in $A_i$ is called an *ordinary local network*. The set of $k$ local networks is called a *generalized similarity network* of $P$. When $A_1, ..., A_k$ is a partition of *domain(H)*, then the set of $k$ comprehensive local networks is called a *generalized Bayesian multinet*.

For example, the secured-building story is represented in the generalized similarity network of Figure 7. Note, $H = \{h_1, h_2\}$ and *domain(H)* consists of nine elements $(x, y)$ where both $x$ and $y$ are drawn from the set $\{w, v, s\}$. A connected cover of *domain(H)* upon which Figure 7 is based consists of: $\{(s,s)\,(v,s)\,(s,v)\,(v,v)\}$, $\{(v,v)\,(w,v)\,(v,w)\,(w,w)\}$, and $\{(s,s)\,(s,w)\,(w,s)\}$. This cover is connected.

Figure 7: A generalized similarity network with two hypothesis nodes.

Most asymmetric independence assertions encoded in Figure 7 were either explained in previous sections or are obvious from the verbal description of the story.

The absence of a link between $h_1$ and $h_2$ in the top network encodes the fact that if the guard knew that one person is a spy, this knowledge would not help him/her decide whether the other person is a spy or a visitor. The existence of a link between $h_1$ and $h_2$ in the middle network encodes the fact that workers come in pairs more often than do visitors. Hence the knowledge that one person is a worker is a clue for classifying the other person.

The vanishing dependence between hypothesis variables $h_1$ and $h_2$ in case of spies vs. visitors is an example of *inter-hypothesis independence*. Such asymmetric assertions cannot be encoded in ordinary similarity networks.

## 5 Summary

This paper proposes an efficient format for encoding and using asymmetric independence assertions for inference. The model builder is asked to express knowledge about independence by constructing multiple local networks using informal guidelines of causation and time ordering. Like any Bayesian network, local networks possess precise semantics in terms of independence assertions and these can be used to verify 1) whether the network faithfully represents the domain and 2) whether the input is consistent.

Multiple local networks have several advantages compared to a single Bayesian network. The elicitation of several small networks is easier than eliciting a single full-scale Bayesian network because the expert can focus his/her attention to particular subdomains, and hence, provide more reliable judgments. Multiple networks represent a domain better because more knowledge about independence is qualitatively encoded. Algorithms for finding the most likely hypothesis run faster when using multiple networks. And finally, the overall storage requirement of multiple networks is often smaller than that of a single Bayesian network because as independence assertions become more detailed, less numeric parameters are needed for describing a domain.

Notably, when independence assertions in the domain are symmetric, a single Bayesian network is preferable.

The challenges remain to 1) devise additional graphical representation schemes of salient patterns of independence assertions, (2) provide computer-aided elicitation procedures for constructing these representations, and (3) devise efficient inference procedures that make use of the encoded assertions.